\documentclass{article}

\usepackage[preprint]{neurips_2019}
\pdfoutput=1 
\usepackage[utf8]{inputenc} % allow utf-8 input
\usepackage[T1]{fontenc}    % use 8-bit T1 fonts
\usepackage{hyperref}       % hyperlinks
\usepackage{url}            % simple URL typesetting
\usepackage{booktabs}       % professional-quality tables
\usepackage{amsfonts}       % blackboard math symbols
\usepackage{nicefrac}       % compact symbols for 1/2, etc.
\usepackage{microtype}      % microtypography
\usepackage[pdftex]{graphicx}

\title{PPO Dash: Improving Generalization in Deep Reinforcement Learning}

\author{%
  Joe Booth \\
  Vidya Gamer, LLC\\
  Seattle, WA\\
  \texttt{joe@joebooth.com} \\
}

\begin{document}

\maketitle

\begin{abstract}
  Deep reinforcement learning is prone to overfitting, and traditional benchmarks such as Atari 2600 benchmark can exacerbate this problem. The Obstacle Tower Challenge addresses this by using randomized environments and separate seeds for training, validation, and test runs. This paper examines various improvements and best practices to the PPO algorithm using the Obstacle Tower Challenge to empirically study their impact with regards to generalization. Our experiments show that the combination provides state-of-the-art performance on the Obstacle Tower Challenge.
\end{abstract}

\section{Introduction}
Solving generalization in Deep Reinforcement Learning is an active area of research and recently there has been a growing recognition that traditional benchmarks, such as Atari can compound this problem. The Obstacle Tower Environment, \cite{Juliani2019ObstacleTower}, and Obstacle Tower Challenge were specifically developed to address the problem of generalization by implementing randomized levels, multiple graphics themes and the ability to hold back seeds so that test runs can be assessed using versions of the environment that the researcher has had no access to.

The PPO algorithm, \cite{Schulman2017ProximalPO}, has demonstrated an ability to scale up to and perform well against large scale generalization problems, \cite{Schulman2017ProximalPO}. In OpenAI Five,  \cite{OpenAI_dota}, the authors remark on their surprise at the PPO algorithm's ability to scale up.

In contrast, the authors of the Obstacle Tower Challenge, \cite{Juliani2019ObstacleTower}, document how the PPO algorithm, \cite{Schulman2017ProximalPO}, did not perform well on the configurations of the Obstacle Tower Challenge, \cite{Juliani2019ObstacleTower}, which demand the most generalization, see Figure \ref{fig:poorPPO}. 

\begin{table}[!h]
  \centering
  \begin{tabular}{lc}
    \textbf{algorithm}           & \textbf{Floor mean (std-div)} \\ \hline
    PPO                          & 0.8 (0.4)                     \\
    Rainbow                      & 3.2 (1.1)                     \\
  \end{tabular}
  \caption{Obstacle Tower Challenge authors found that PPO did not generalize well.}
  \label{fig:poorPPO}
\end{table}

So, is the PPO algorithm a good choice for addressing generalization problems? We would like to resolve this question.

In this paper, we introduce PPO-Dash, a set of improvements and best practices to the PPO algorithm and demonstrate state of the art performance on the Obstacle Tower Challenge, \cite{Juliani2019ObstacleTower}. We study the individual elements of PPO-Dash in an attempt to better understand their value.

Our goal with this work is to focus on the generalization aspect of the Obstacle Tower Challenge, \cite{Juliani2019ObstacleTower}, and create a baseline from which we can implement more novel techniques to study the sparse rewards aspects of the environment.

\subsection{The Obstacle Tower Environment}
\begin{figure}[!h]
  \begin{center}
    \includegraphics[width=0.8\linewidth]{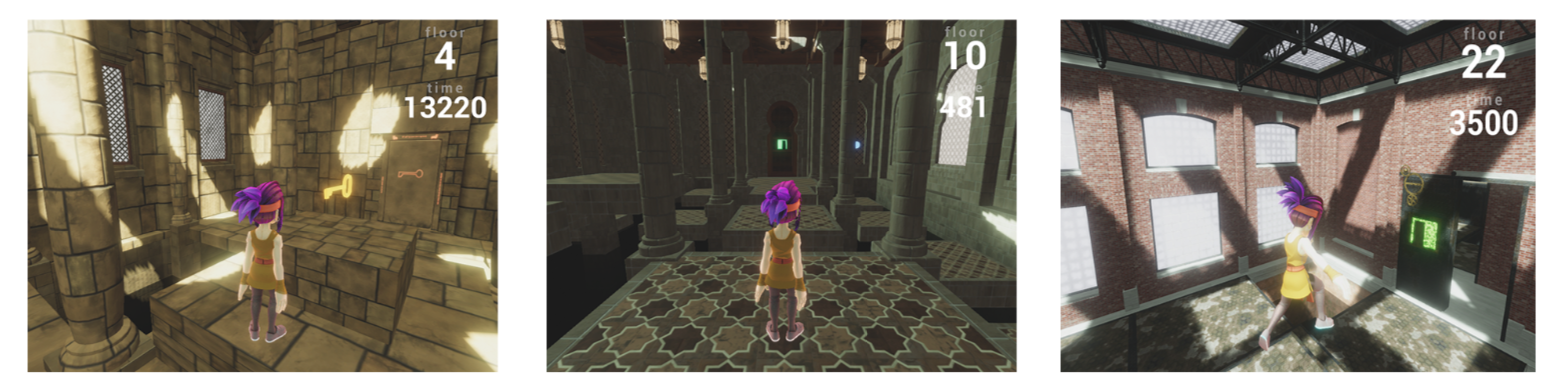}
    \caption{The Obstacle Tower Challenge. Showing three different visual themes.}
    \label{fig:validationPerformance}
  \end{center}
\end{figure}

The Obstacle Tower Environment is a 3D platform game viewed from the third person. Each floor has a set of rooms and puzzles that incrementally increase in difficulty. The environment can be controlled via a human or via a policy using the OpenAI Gym \cite{Brockman2016OpenAIG} interface. 

The player character must solve each room and find the exit of the floor. They then progress to the next floor. If they run out of time or 'die', the tower is reset and they must begin back at floor 0.

The layout and visual theme of each floor are set using a seed. The environment has a total of 25 floors, three visual themes and 100 seeds which randomize the level layout and layout of puzzles. 

The environment exposes pixel observation and a vector observation of the remaining time and number of keys. These can be combined in 'retro' mode whereby the vector observations are visually embedded in the pixels. This is to support older algorithms which only take pixel inputs.

For the purpose of this paper, we use version 1.3 of the environment.

\subsection{The Obstacle Tower Challenge 2019 competition}
Unity has partnered with Google Cloud and AI Crowd to host a competition. The competition consists of two rounds. Round one ran from Monday, February 11th, 2019, through to Tuesday, April 30th, 2019, and is based on version one of the Obstacle Tower Environment. This environment has a total of 25 floors, three visual themes and 100 seeds which randomize the level layout.

Round two runs from Wednesday, May 15th, 2019 to Monday, July 15th, 2019 and is based on version 2 of the Obstacle Tower Environment. This environment has a total of 100 floors and it adds two additional visual themes, new puzzles, and enemy types.

The competition is hosted on the AI Crowd platform. Submissions are made by submitting a repository containing the trained model and code needed to run the model. AI Crowd performs the test using a special version of the Obstacle Tower Environment which contains five seeds which are not available in the public version. One can view the results (the breakdown of how the policy performed against each seed) via the AI Crowd website. The final result is submitted to the leaderboard for the current round. The last entry per user is the one displayed on the leaderboard.

\subsection{Reproducibility}
We have completed two full training passes of 200,000,000 training steps each. The validation score is calculated using seeds held back from the training set of seeds. The test scores (the official round one leaderboard scores) were generated by AI Crowd / Unity (the competition organizers) using a special version of the Obstacle Tower Environment that exposes five unique seeds.

We used a consumer grade desktop with an i7 8700k, 32GB of ram and a GTX 1080 GPU. Training takes approximately seven hrs for 10,000,000 steps (140 hours for 200m steps).

All source code, trained models, raw Tensorboard files and instructions can be found here: 
https://github.com/Sohojoe/ppo-dash

\section{PPO Dash}

Our general approach for PPO Dash was to look for improvements and best practices which would a) reduce the action and observation space, b) be appropriate for 3D exploration environments, and c) leverage general best practices with the PPO algorithm.

\subsection{Training and testing methodology}
We train in runs of 10m, 20m or 50m steps. We held back five seeds (95 - 99) during early training to enable a faster validation on unseen seeds. Our validation pass runs five times across each seed and gives the average floor of all runs, we did this as we experienced a wide set of results on some seeds and so averaging over five runs gives us a better indication of the underlying quality of our policy. We also ran validation runs on the five seeds that unity provided for validation (seeds 100-105). 

After 100m training steps we reduce the epoch to one and the learning rate to 1-e4. This was to help the policy learn more subtle aspects of the environment.

The first version of our agent (PPO Dash - no recurrent) did not have recurrent memory which we added for the second version of our agent (PPO Dash).

\subsection{Implementation Details}
We based our code on an existing implementation of PPO, \cite{pytorchrl} and made the following changes:

\begin{itemize}
  \item \textbf{Action Space Reduction} The goal of Action Space Reduction is to increase learning performance by reducing the number of required network outputs. We chose to be aggressive in this task because we wanted the network to focus on the greater generalization goal and not to be burdened with learning the semantic relationship between actions which a human takes for granted.
  \begin{itemize}
    \item We chose a set of 8 actions (a drop of 85\% from the original 54 actions). 
    \item These actions are: no-action, forward, rotate left, rotate right, jump+forward, backward, forward + rotate left, forward+ rotate right. 
    \item We playtested these actions to ensure it was possible for a human to progress. 
    \item The rationale for reducing jump actions to the single jump + forward action is because once the jump has begun, other actions such as no-op/forward/backward and rotate camera will still direct the jump with minimal loss of control.
    \item The rationale for reducing backward actions to a single backward action is because most actions are forward based. 
    \item The rationale for reducing forward actions to a single forward action is because the actor can stop, rotate left and right, then move in the new direction. 
  \end{itemize}
  \item \textbf{Frame Stack Reduction} The goal of Frame Stack Reduction is to increase learning performance by reducing the number of required network inputs. 
  \begin{itemize}
    \item \cite{Mnih2015HumanlevelCT} noted that the Atari 2600 had a limited number of hardware sprites as many Atari games use a technique of alternating sets of sprites between frames. Thus, \cite{Mnih2015HumanlevelCT} introduced the practice of stacking the last four frames as observations. We hypothesized that this practice is not needed for most environments and that the frame stack can be reduced to two frames, or even one. 
    \item We experimented between two and one historical frames; we found no noticeable difference and so chose a single frame to maximize the reduction of the observation space.
    \item We further reduced the observation by only stacking the brightness of the historical frame.
  \end{itemize}
  \item \textbf{Large Scale Hyperparameters} Typically a PPO algorithm's hyperparameters are tuned for the Atari benchmark. Our hypothesis was that the Obstacle Tower Challenge environment was closer to the Unity Maze environment in \cite{Burda2019LargeScaleSO} as both environments are 3D in nature and both environments have sparse rewards. We implemented the following changes to the hyperparameters:
  \begin{itemize}
    \item 8 epochs
    \item Learning Rate of 1e-4 at a constant learning rate (no linear decay)
    \item Entropy-coef = 0.001
    \item Number of mini-batches = 8 
    \item Number of concurrent agents = 32  
    \item Number of steps per epoch = 512 - \cite{Burda2019LargeScaleSO} authors note that their rationale for 512 steps and 8 epochs was so that the policy can lock on to the sparse reward.
  \end{itemize}
  \item \textbf{Vector Observations} The goal of adding Vector Observations was to maximize the use of visible pixels and to remove the burden of the policy needing to learn visually encoded state information. 
  \begin{itemize}
    \item Typically, benchmarks such as Atari only expose visual observations and game state. To support existing algorithms Obstacle Tower Challenge provides a 'retro mode' whereby the health and inventory status is encoded into the top few lines of the screen. This has the disadvantage of reducing the total visible space that the agent observers.
    \item The Obstacle Tower Challenge exposes vectorized observations of the current health and of the count of the current number of keys (which we one-hot encode)
  \end{itemize}
  \item \textbf{Normalized Observations} The goal of normalizing observations is to help increase the variance per pixel. We build a per pixel mean and global standard deviation by running the environment using randomized steps for 10,000 steps over a randomized set of seeds and floors.
  \item \textbf{Reward Hacking} The goal of reward hacking is to improve the reward signal. By default, Obstacle Tower Challenge gives a reward of 1 for completing the floor and a reward of 0.1 for completing a puzzle such as opening a door, finding a key, pushing the block on to a target. We made the following changes:
  \begin{itemize}
    \item Completing the floor: we add the remaining health to encourage the policy to finish each floor as quickly as possible (the range of reward is between 1 and 4)
    \item puzzle completion: We give a reward of 1 (by default this is 0.1)
    \item Health pickup: We give a reward of 0.1 for picking up a blue dot/health pick up.
    \item On game over: When running out of time, we give a reward of -1 
  \end{itemize}
  \item \textbf{Recurrent Memory} Our motivation for adding recurrent memory was that it would help the policy learn temporal aspects of solving puzzles. We used the existing implementation of recurrent memory from \cite{pytorchrl}
\end{itemize}

\subsection{Observations and Results}
Figure \ref{fig:validationPerformance} and Table \ref{validationPerformance2} shows the performance of the PPO-Dash agent on the validation seeds (100-105). The agent is not able to complete floor 10. However, it does learn to complete the other seeds and this generalizes to the validations seeds. 

\begin{figure}[!h]
  \begin{center}
    \includegraphics[width=0.8\linewidth]{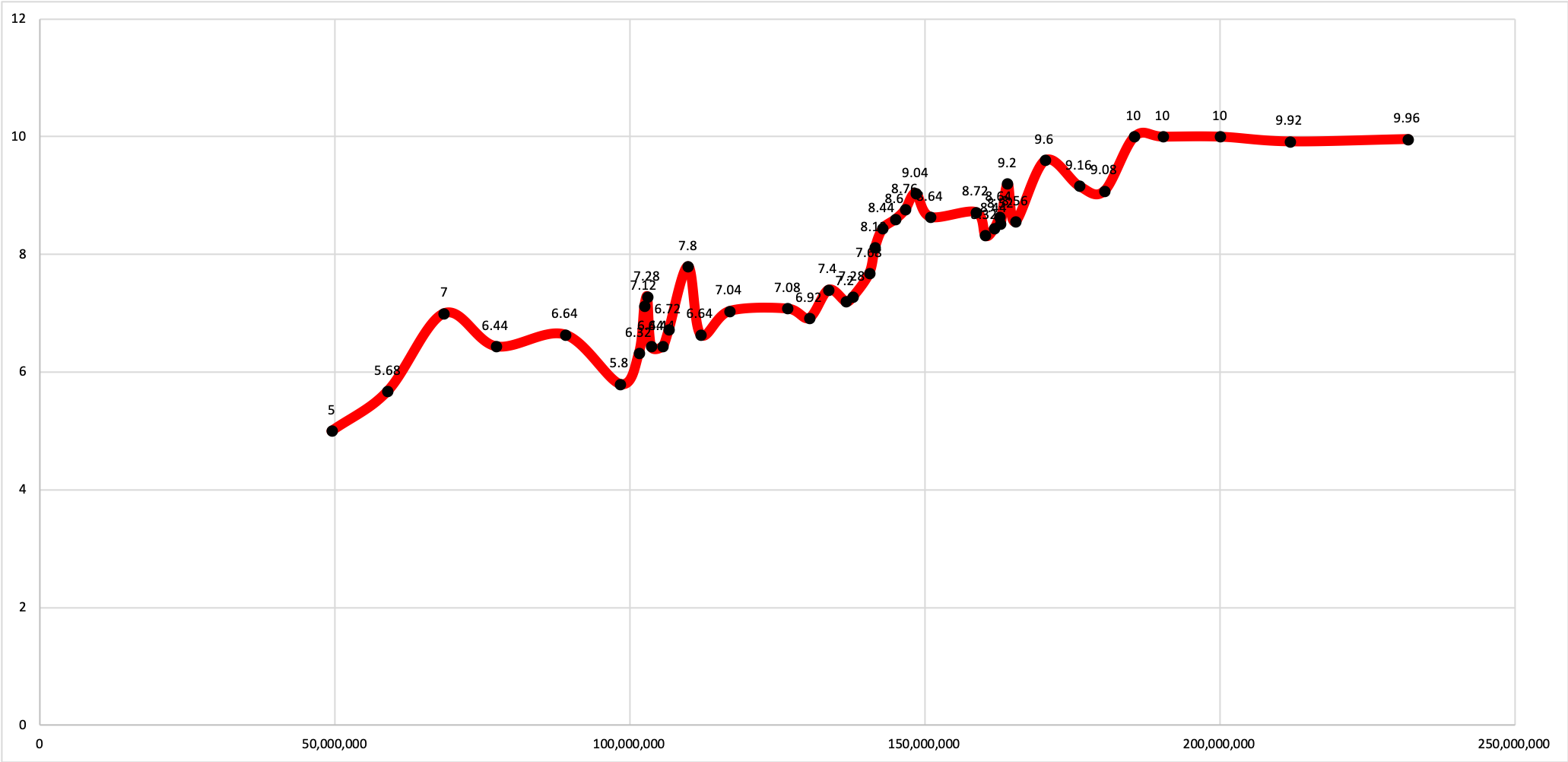}
    \caption{PPO Dash Validation Performance. The agent scores against the validation seeds. The dots indicate a validation run (with score). The horizontal axis is the number of training steps, and the vertical axis is the mean floor.}
    \label{fig:validationPerformance}
  \end{center}
\end{figure}

\begin{table}[!h]
  \centering
  \begin{tabular}{lc}
    \textbf{algorithm}           & \textbf{Floor mean (std-div)} \\ \hline
    PPO                          & 0.8 (0.4)                     \\
    Rainbow                      & 3.2 (1.1)                     \\
    \textbf{PPO Dash (ours)}     & \textbf{10 (0.0)}             \\
    PPO Dash no recurrent (ours) & 9.2 (0.97)                   
  \end{tabular}
  \caption{Results vs prior state of art}
  \label{validationPerformance2}
\end{table}

In Table \ref{fig:round1}, we see the performance of the PPO-Dash agent against the test seeds as evaluated by Unity and their partners. These are the official standings from the end of round one which show that PPO-Dash achieved 2nd place\footnote{https://blogs.unity3d.com/2019/05/15/obstacle-tower-challenge-round-2-begins-today/}.

\begin{table}[!h]
  \centering
  \begin{tabular}{lccc}
    \textbf{Participant}     & \textbf{Round 1 Average Floors} & \textbf{Round 1 Average Reward} & \textbf{\begin{tabular}[c]{@{}c@{}}Final Position\\ (Round 1)\end{tabular}} \\ \hline
    unixpickle               & \textbf{16.40}                  & 29.88                           & 1                                                                           \\
    \textbf{ppo-dash (ours)} & 10.00                           & 16.46                           & 2                                                                           \\
    dougm                    & 9.61                            & 15.92                           & 3                                                                           \\
    karolisram               & 8.40                            & 13.32                           & 4                                                                           \\
    sova876                  & 8.20                            & 13.12                           & 5                                                                           \\
    giadefa                  & 8.00                            & 12.82                           & 6                                                                           \\
    wywarren                 & 8.00                            & 12.58                           & 7                                                                           \\
    PerInDisguise            & 7.00                            & 10.68                           & 8                                                                           \\
    tatsuyaogawa             & 6.60                            & 9.72                            & 9                                                                           \\
    STAR.Lab                 & 6.00                            & 8.72                            & 10                                                                         
  \end{tabular}
  \caption{Offical results from round one of the Obstacle Tower Challenge. PPO-Dash averaged floor 10, finishing second.}
  \label{fig:round1}
\end{table}

\section{Study of how individual elements impact learning}

We would like to investigate how each element of PPO-Dash impacts learning. To gain this insight we use two strategies. The first strategy is to compare individual elements against a baseline. The second strategy is to incrementally add PPO-Dash elements and compare the incremental impact.

\subsection{Method}

To create our baseline we look at \cite{Juliani2019ObstacleTower} and reproduce their results. For the most part, this is using the default hyperparameters from the PPO implementation from OpenAI.Baselines \cite{baselines} with 50 concurrent agents.

\subsubsection{Compare elements against the baseline}
The elements we compare against the baseline are:
\begin{itemize}
  \item Reduce Action Space
  \item Large Scale Hyperparameters
  \item Reduced Frame Stack
  \item Recurrent
\end{itemize}

\subsubsection{Compare elements incrementally}
The elements we incrementally combine are:
\begin{itemize}
  \item Reduced Frame Stack
  \item Reduced Frame Stack + Reduced Action Space
  \item Reduced Frame Stack + Reduced Action Space + Large Scale Hyperparameters
  \item Reduced Frame Stack + Reduced Action Space + Large Scale Hyperparameters + Recurrent
  \item Reduced Frame Stack + Reduced Action Space + Large Scale Hyperparameters + Recurrent + Vector Observations
\end{itemize}

\subsubsection{Compare different sets of actions}
We experimented with several groups of action sets. We asked a human to playtest each action set to confirm that they could progress through to level 10. We validated our choice by comparing each group of action sets over 10m steps.

The action sets we tested where
\begin{itemize}
%  \item Action Set 5 - no-action, forward, rotate left, rotate right, jump+forward
  \item Action Set 6 - no-action, forward, rotate left, rotate right, jump+forward, backward
  \item Action Set 8 - no-action, forward, rotate left, rotate right, jump+forward, backward, forward + rotate left, forward+ rotate right
  \item Action Set 20 - all combinations of no-action, forward, left, right, rotation left, rotate right. A single backward action (backward) and single jump action (forward and jump) 
  \item Action Set 27 - all combinations of no-action, forward, backward, left, right, rotation left, rotate right. A single jump action (forward and jump)
  \item Action Set 54 - all actions
\end{itemize}

\subsubsection{Reproducibility}

For the empirical study of comparing elements with the baseline and comparing elements incrementally, we used three training runs. We record the mean floor and standard deviation at each timestep and average these over the three runs.

When comparing the different sets of actions we only ran each action set once.

We use a consumer grade desktop with an i7 8700k, 32GB of ram and an RTX 2080 Ti GPU. Each individual training run takes approximately five hours.

All source code, trained models, raw tensorboard files and instructions can be found here: 
https://github.com/Sohojoe/ppo-dash

\subsection{Results}

We found that, with the exception of Reduced Action Space, our results were inconclusive. We believe that this was because we limited training to 10m steps. In addition to the reported experiments, we did try some alternative hyperparameters which did not show a change to this trend.

\subsubsection{Results comparing elements against the baseline}
When comparing the results of individual elements against the baseline, we found that Reduced Action Space showed a significant improvement over the baseline (mean of 4.49 vs mean of 1.21). However, all the other comparables where very close to the baseline, see Figure \ref{fig:compare_baseline}

\begin{table}[!h]
  \centering
  \begin{tabular}{lc}
    & Mean \\
    Baseline                    & 1.21 \\ \hline
    Reduced Action Space        & 4.49 \\
    Large Scale Hyperparameters & 1.94 \\
    Reduced Frame Stack         & 1.24 \\
    Recurrent                   & 1.11
  \end{tabular}
  \caption{Comparison of PPO Dash elements compared against a baseline implementation of PPO. Mean floor over three runs.}
  \label{fig:compare_baseline}  
  \end{table}

\subsubsection{Results comparing elements incrementally}
The results stacking PPO-Dash elements incrementally are inconclusive. We see a significant jump in the mean floor when adding Reduced Action Space, however, other elements do not have much of an impact. See Figure \ref{fig:validationPerformance}

\begin{figure}[!h]
  \begin{center}
    \includegraphics[width=0.8\linewidth]{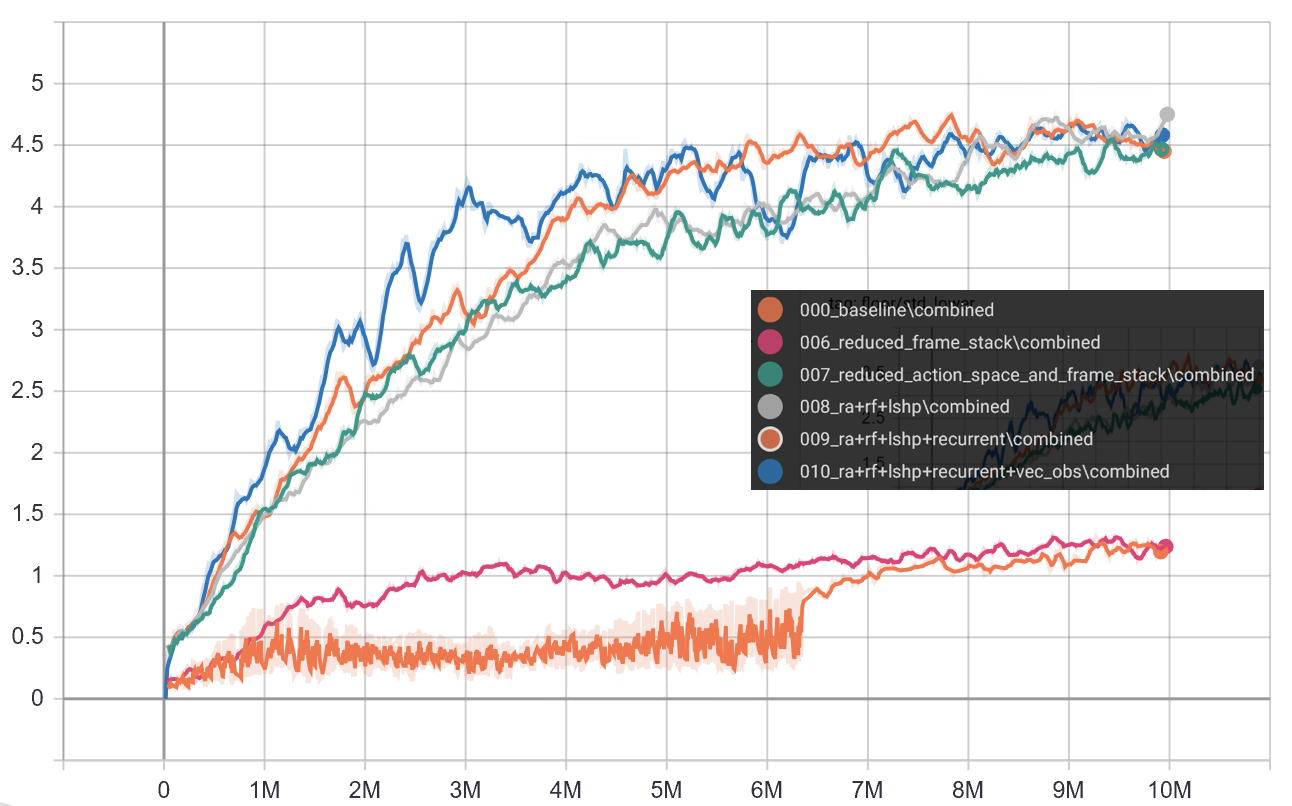}
    \caption{Results comparing PPO-Dash elements incrementally. Mean floor over three runs of 10m steps each.}
    \label{fig:validationPerformance}
  \end{center}
\end{figure}

\subsubsection{Results on comparing different sets of actions}
We found that the 8, 20, and 27 action sets all saw a similar improvement in performance going from an average of three floors to an average of five floors (see Table \ref{fig:action_sets}).  

\begin{table}[!h]
  \centering
  \begin{tabular}{lccccc}
    \hline
                           & \begin{tabular}[c]{@{}c@{}}Action\\ Set 6\end{tabular} & \begin{tabular}[c]{@{}c@{}}Action\\ Set 8\end{tabular} & \begin{tabular}[c]{@{}c@{}}Action\\ Set 20\end{tabular} & \begin{tabular}[c]{@{}c@{}}Action\\ Set 27\end{tabular} & \begin{tabular}[c]{@{}c@{}}Action\\ Set 54\end{tabular} \\ \hline
    Limit Jump Actions     & \checkmark                                              & \checkmark                                              & \checkmark                                               & \checkmark                                               &                                                         \\
    Limit Backward Actions & \checkmark                          & \checkmark                          & \checkmark                           &                                     &                                     \\
    Limit Forward Actions  & \checkmark                                              &                                                        &                                                         &                                                         &                                                         \\\hline
    \textbf{Ave Floor}     & 3                                                      & \textbf{5}                                             & \textbf{5}                                              & \textbf{5}                                              & 3                                                       \\ \hline
  \end{tabular}
  \caption{Comparison of action sets and their impact on the maximum floor reached after 10m steps.}
  \label{fig:action_sets}  
\end{table}

\section{Conclusion}

We introduce PPO-Dash, a collection of various improvements and best practices to the PPO algorithm that produced state of the art on the Obstacle Tower Challenge. Within the content of the Obstacle Tower Environment and the associated Obstacle Tower Challenge (competition), we demonstrated generalization from the training seeds, through validation seeds, through to test seeds. 

These results exceeded our expectations of how well an algorithm could perform against this task without specifically addressing sparse reward. They give strong evidence that PPO does scale for these types of tasks.

We studied the impact of individual elements of PPO-Dash against a baseline and by stacking them together. Our study proved to be inconclusive shedding little insight into the impact of most of the elements. This finding has value in that it gives evidence that to show insight against this challenge, potentially many more steps are needed. For example, it may be the case that 150-200m steps are needed, a 15-20x multiple on what we used, which would take 75 to 100 hours per run on the hardware we used.

Our study of different sized action sets indicated that there is a big gain in performance by reducing the jump action and then little variance between the 8, 20, 27 sized sets and limiting the forward set of actions (the 6 set), did impact performance. We stress that these are early indicators both because we did not conform to reproducibility best practices (for this study) and because of the insight we gained with regards to the limit of only running over 10m steps.

In terms of further work, it would be valuable to find the number of steps needed to do reliable feature studies and also how these best practices translate to other reinforcement algorithms, for example, Rainbow-Dash. 

\bibliography{general} 
\bibliographystyle{apalike}

\end{document}